\documentclass[10pt,twocolumn,letterpaper]{article}

\usepackage{cvpr}
\usepackage{times}
\usepackage{epsfig}
\usepackage{graphicx}
\usepackage{amsmath}
\usepackage{amssymb}
\usepackage[hyphens]{url}

\usepackage[breaklinks=true,bookmarks=false]{hyperref}

\cvprfinalcopy % *** Uncomment this line for the final submission

\def\cvprPaperID{****} % *** Enter the CVPR Paper ID here

\begin{document}

%%%%%%%%% TITLE
\title{Attention Branch Network:\\ Learning of Attention Mechanism for Visual Explanation}

%\author{Hiroshi Fukui\\
%Chubu university\\
%1200 Matsumotocho, Kasugai, Aichi, Japan\\
%{\tt\small fhiro@mprg.cs.chubu.ac.jp}
%\and
%Tsubasa Hirakawa\\
%{\tt\small hirakawa@mprg.cs.chubu.ac.jp}
%\and
%Takayoshi Yamashita\\
%{\tt\small yamashita@isc.chubu.ac.jp}
%\and
%Hironobu Fujiyoshi\\
%{\tt\small fujiyoshi@isc.chubu.ac.jp}
%}

\author{Hiroshi Fukui, Tsubasa Hirakawa, Takayoshi Yamashita, Hironobu Fujiyoshi\\
Chubu University\\
1200 Matsumotocho, Kasugai, Aichi, Japan\\
{\tt\small \{fhiro@mprg.cs, hirakawa@mprg.cs, yamashita@isc, fujiyoshi@isc\}.chubu.ac.jp}
}


\maketitle
%\thispagestyle{empty}

%%%%%%%%% ABSTRACT
\begin{abstract}
Visual explanation enables humans to understand the decision making of deep convolutional neural network~(CNN), but it is insufficient to contribute to improving CNN performance.
In this paper, we focus on the attention map for visual explanation, which represents a high response value as the attention location in image recognition.
This attention region significantly improves the performance of CNN by introducing an attention mechanism that focuses on a specific region in an image.
In this work, we propose Attention Branch Network~(ABN), which extends a response-based visual explanation model by introducing a branch structure with an attention mechanism.
ABN can be applicable to several image recognition tasks by introducing a branch for the attention mechanism and is trainable for visual explanation and image recognition in an end-to-end manner. 
We evaluate ABN on several image recognition tasks such as image classification, fine-grained recognition, and multiple facial attribute recognition.
Experimental results indicate that ABN outperforms the baseline models on these image recognition tasks while generating an attention map for visual explanation.
Our code is available \footnote{\url{https://github.com/machine-perception-robotics}\\ \url{-group/attention_branch_network}}.
\end{abstract}

%%%%%%%%% BODY TEXT
\section{Introduction}
Deep convolutional neural network~(CNN)~\cite{Alex2014,LeCun1989} models have been achieved the great performance on various image recognition tasks~\cite{Ren2015,He2016,Emily2017,Xie2017,He2017,Hu2017,Chen2018}.
However, despite CNN models performing well on such tasks, it is difficult to interpret the decision making of CNN in the inference process.
To understand the decision making of CNN, methods of interpreting CNN have been proposed~\cite{Zeiler2014,Zhou2016,Ribeiro2016,Smilkov2017,Selvaraju2017,Aditya2017,Montavon2018}.

\begin{figure}[t]
\begin{center}
\includegraphics[width=0.95\linewidth]{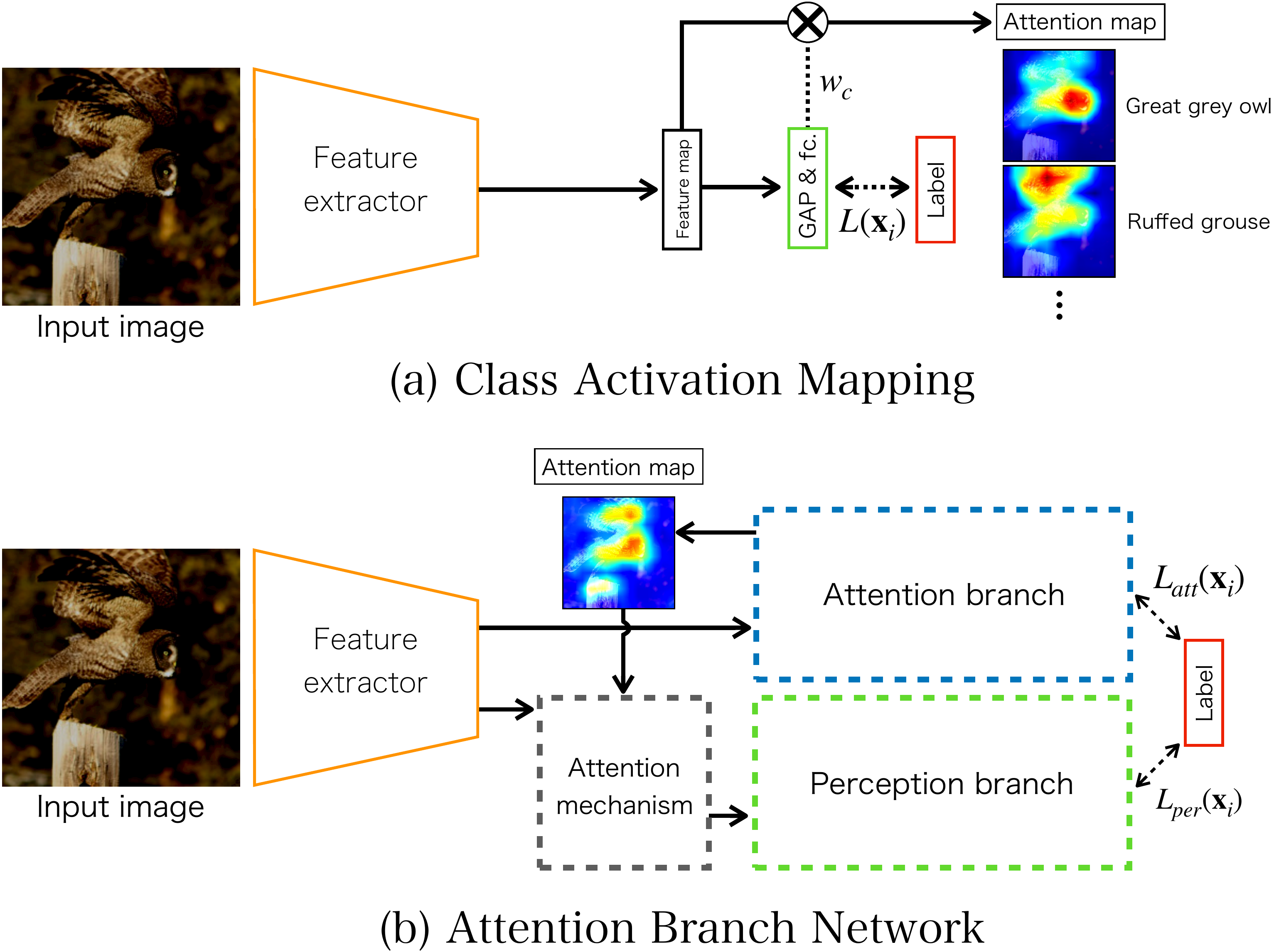}
\caption{Network structures of class activation mapping and proposed attention branch network.}
\label{fig:Intro_abn}
\end{center}
\end{figure}

``Visual explanation" has been used to interpret the decision making of CNN by highlighting the attention location in a top-down manner during the inference process.
Visual explanation can be categorized into gradient-based or response-based.
Gradient-based visual explanation typically use gradients with auxiliary data, such as noise~\cite{Smilkov2017} and class index~\cite{Selvaraju2017,Aditya2017}.
Although these methods can interpret the decision making of CNN without re-training and modifying the architecture, they require the backpropagation process to obtain gradients.
In contrast, response-based visual explanation can interpret the decision making of CNN during the inference process.
Class activation mapping~(CAM)~\cite{Zhou2016}, which is a representative response-based visual explanation, can obtain an attention map in each category using the response of the convolution layer. 
CAM replaces the convolution and global average pooling~(GAP)~\cite{Lin2014} and obtains an attention map that include high response value positions representing the class, as shown in Fig.~\ref{fig:Intro_abn}(a).
However, CAM requires replacing the fully-connected layer with a convolution layer and GAP, thus, decreasing the performance of CNN.

To avoid this problem, gradient-based methods are often used for interpreting the CNN.
The highlight location in an attention map for visual explanation is considered an attention location in image recognition.
To use response-based visual explanation that can visualize an attention map during a forward pass, we extended a response-based visual explanation model to an attention mechanism.
By using the attention map for visual explanation as an attention mechanism, our network is trained while focusing on the important location in image recognition.
The attention mechanism with a response-based visual explanation model can simultaneously interpret the decision making of CNN and improve their performance.

Inspired by response-based visual explanation and attention mechanisms, we propose $Attention$ $Branch$ $Network$~(ABN), which extends a response-based visual explanation model by introducing a branch structure with an attention mechanism, as shown in Fig~\ref{fig:Intro_abn}(b).
ABN consists of three components: feature extractor, attention branch, and perception branch.
The feature extractor contains multiple convolution layers for extracting feature maps.
The attention branch is designed to apply an attention mechanism by introducing a response-based visual explanation model.
This component is important in ABN because it generates an attention map for the attention mechanism and visual explanation.
The perception branch outputs the probabilities of class by using the feature and attention maps to the convolution layers.
ABN has a simple structure and is trainable in an end-to-end manner using training losses at both branches.
Moreover, by introducing the attention branch to various baseline model such as ResNet~\cite{He2016}, ResNeXt~\cite{Xie2017}, and multi-task learning~\cite{Caruana1993}, ABN can be applied to several networks and image recognition tasks.

Our contributions are as follows:
\begin{itemize}
\item ABN is designed to extend a response-based visual explanation model by introducing a branch structure with an attention mechanism.
ABN is the first attempt to improve the performance of CNN by including a visual explanation.
\item ABN is applicable to various baseline models such as VGGNet~\cite{Simonyan2014}, ResNet~\cite{He2016}, and multi-task learning~\cite{Caruana1993} by dividing a baseline model and introducing an attention branch for generalizing an attention map.
\item By extending the attention map for visual explanation to attention mechanism, ABN simultaneously improves the performance of CNN and visualizes an attention map during forward propagation.
\end{itemize}

\section{Related work}

\subsection{Interpreting CNN}
Several visual explanation, which highlight the attention location in the inference process, have been proposed~\cite{Van2008,Zeiler2014,Zhou2016,Ribeiro2016,Springenberg2015,Smilkov2017,Selvaraju2017,Aditya2017,Montavon2018}.
There two types of visual explanation: gradient-based visual explanation, which uses a gradient and feed forward response to obtain an attention map, and response-based visual explanation, which only uses the response of a feed forward propagation.
With gradient-based visual explanation, SmoothGrad~\cite{Selvaraju2017} obtains sensitivity maps by adding noise to the input image iteratively and takes the average of these sensitivity maps.
Guided backpropagation~\cite{Springenberg2015} and gradient-weighted class activation mapping~(Grad-CAM)~\cite{Smilkov2017,Aditya2017}, which are gradient-based visual explanation, have been proposed.
Guided backpropagation and Grad-CAM visualize an attention map using positive gradients  at a specific class in backpropagation.
Grad-CAM and guided backpropagation have been widely used because they can interpret various pre-trained models using the attention map of a specific class.

Response-based visual explanation visualizes an attention map using the feed forward response value from a convolution or deconvolution layer.
While such models require re-training and modifying a network model, they can directly visualize an attention map during forward pass.
CAM~\cite{Zhou2016} can visualize an attention maps for each class using the response of a convolution layer and the weight at the last fully-connected layer.
CAM performs well on weakly supervised object localization but not as well in image classification due to replacing fully-connected layers with convolution layers and passing through GAP.

We constract ABN by extending the CAM, which can visualize an attention map for visual explanation in feed forward propagation, to an attention mechanism.
CAM is easily compatibles with an attention mechanism that directly weights the feature map.
In contrast, gradient-based visual explanation is not compatible with ABN because it requires the back propagation process to obtain the gradients.
Therefore, we use CAM as the attention mechanism for ABN.

\begin{figure*}[htbp]
\begin{center}
\includegraphics[width=0.92\linewidth]{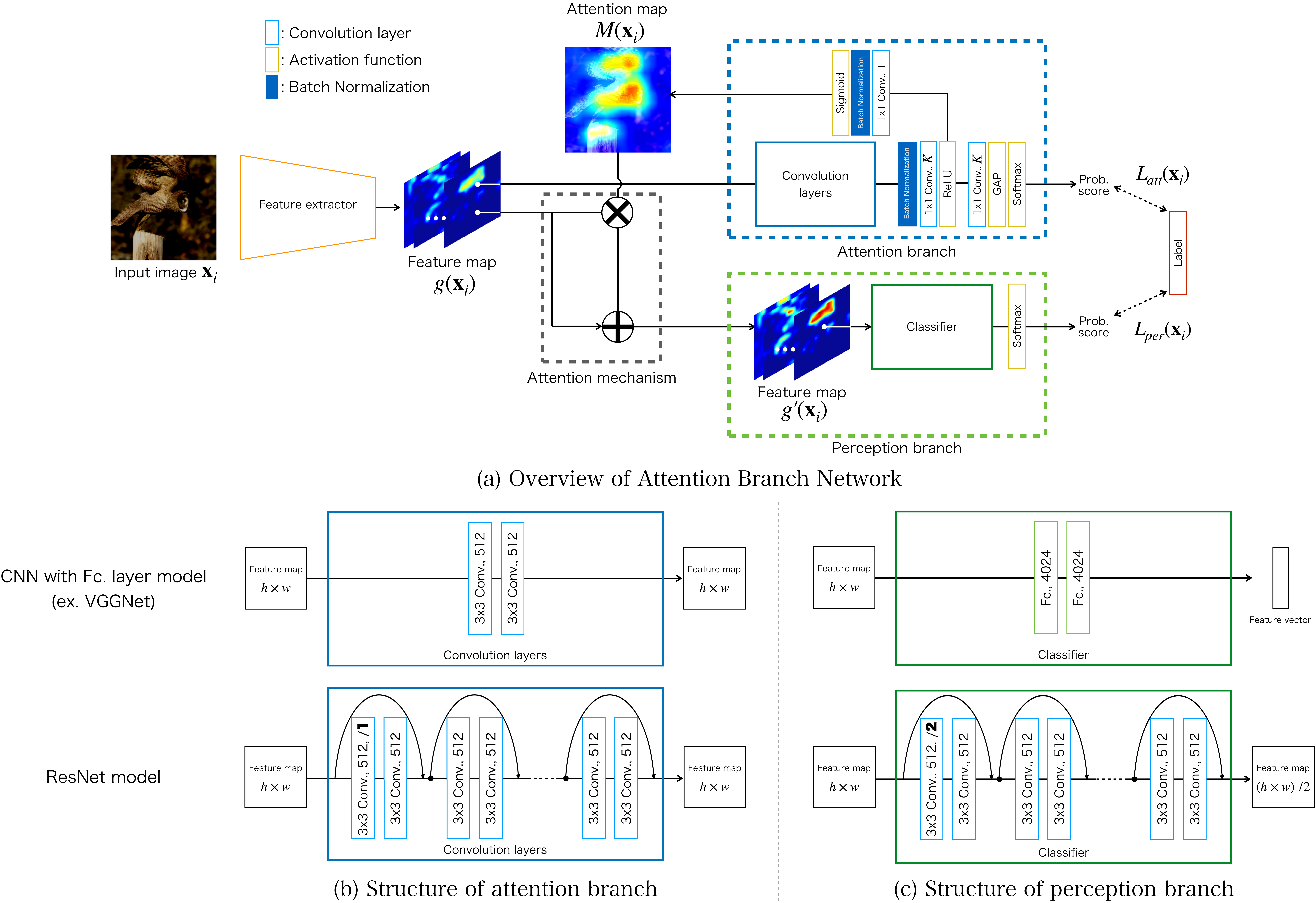}
\caption{Detailed structure of Attention Branch Network.}
\label{fig:attebtion_branch_network}
\end{center}
\end{figure*}

\subsection{Attention mechanism}
Attention mechanisms have been used in computer vision and natural language processing~\cite{Thang2015,Xu2015,Wang2017a,Hu2017}.
They have been widely used in sequential models~\cite{Xu2015,Zichao2016,You2016,Bahdanau2016,Ashish2017} with recurrent neural networks and long short term memory~(LSTM)~\cite{Sepp1997}.
A typical attention model on sequential data has been proposed by Xu $et~al.$~\cite{Xu2015}.
The attention mechanism of their model is based on two types of attention mechanisms: soft and hard.
The soft attention mechanism of Xu $et~al.$ model is used as the gate of LSTM, and image captioning and visual question answering have been used~\cite{Zichao2016,You2016}.
Additionally, the non-local neural network~\cite{Wang2018}, which uses the self-attention approach, and the recurrent attention model~\cite{Volodymyr2014}, which controls the attention location by reinforcement learning, have been proposed.

The recent attention mechanism is also applied to single image recognition tasks~\cite{Wang2017a,Hu2017,Linsley2018}.
Typical attention models on a single image are residual attention network~\cite{Wang2017a} and squeeze-and-excitation network~(SENet)~\cite{Hu2017}.
The residual attention network includes two attention components, i.e., a stacked network structure that consists of multiple attention components, and attention residual learning that applies residual learning~\cite{He2016} to an attention mechanism.
SENet includes a squeeze-and-excitation block that contains a channel-wise attention mechanism introduced for each residual block.

ABN is designed to focus on the attention map for visual explanation that represents the important region in image recognition.
Previous attention models extract a weight for an attention mechanism using only the response value of the convolution layers during feed forward propagation in an unsupervised learning manner.
However, ABN easily extracts the effective weight for an attention mechanism in image recognition by generating the attention map for visual explanation on the basis of response-based visual explanation in a supervised learning manner.

\section{Attention Branch Network}
As mentioned above, ABN consists of three modules: feature extractor, attention branch, and perception branch, as shown in Fig.~\ref{fig:Intro_abn}.
The feature extractor contains multiple convolution layers and extracts feature maps from an input image.
The attention branch outputs the attention location based on CAM to an attention map by using an attention mechanism.
The perception branch outputs the probability of each class by receiving the feature map from the feature extractor and attention map.

ABN is based on a baseline model such as VGGNet~\cite{Simonyan2014} and ResNet~\cite{He2016}.
The feature extractor and perception branch are constructed by dividing a baseline model between a specific layer.
The attention branch is constructed after feature extractor on the basis of the CAM.
ABN can be applied to several image recognition tasks by introducing the attention branch.
We provide ABN for the several image recognition tasks such as image classification, fine-grained recognition, and multi-task learning.

\subsection{Attention branch}
CAM has a $K \times 3 \times 3$ convolution layer, GAP, and, fully-connected layer as last the three layers, as shown in Fig.~\ref{fig:Intro_abn}(a).
Here, $K$ is the number of categories, and ``$K \times 3 \times 3$ convolution layer" means a $3 \times 3$ kernel with $K$ channels at the convolution layer.
The $K \times 3 \times 3$ convolution layer outputs a $K \times h \times w$ feature map, which represents the attention location for each class.
The $K \times h \times w$ feature map is down-sampled to a $1 \times 1$ feature map by GAP and outputs the probability of each class by passing through the fully-connected layer with the softmax function.
When CAM visualizes the attention map of each class, an attention map is generated by multiplying the weighted sum of the $K \times h \times w$ feature map by the weight at the last fully-connected layer.

CAM replaces fully-connected layers with $3 \times 3$ convolution layers.
This restriction is also introduced into the attention branch.
The fully-connected layer that connects a unit with all units at the next layer negates the ability to localize the attention area in the convolution layer.
Therefore, if a baseline model contains a fully-connected layer, such as VGGNet, the attention branch replaces that fully-connected layer with a $3 \times 3$ convolution layer, similar with CAM, as shown at the top of Fig.~\ref{fig:attebtion_branch_network}(b) .
ResNet models with ABN are constructed from the residual block at the attention branch, as shown at the bottom of Fig.~\ref{fig:attebtion_branch_network}(b).
We set the stride of the first convolution layer at the residual block as 1 to maintain the resolution of the feature map.

To generate an attention map, the attention branch builds a top layer based on CAM, which consists of a convolution layer and GAP.
However, CAM cannot generate an attention map in the training process because the attention map is generated using the feature map and weight at a fully-connected layer after training.
To address this issue, we replace the fully-connected layer with a $K\times1\times1$ convolution layer, as with CAM.
This $K\times1\times1$ convolution layer is imitated at the last fully-connected layer of CAM in a feed forward processing.
After the $K\times1\times1$ convolution layer, the attention branch outputs the class probability by using the response of GAP with the softmax function.
Finally, the attention branch generates an attention map from the $K \times h \times w$ feature map.
Then, to aggregate the $K$ feature maps, these feature maps are convoluted by a $1\times1\times1$ convolution layer.
By convoluting with a $1\times1\times1$ kernel, $1\times h \times w$ feature map is generated.
We use the $1\times h \times w$ feature map normalized by the sigmoid function as the attention map for the attention mechanism.

\subsection{Perception branch}\label{sec:perception}
The perception branch outputs the final probability of each class by receiving the attention and feature maps from the feature extractor.
The structure of the perception branch is the same for conventional top layers from image classification models such as VGGNet and ResNet, as shown in Fig.~\ref{fig:attebtion_branch_network}(c).
First, the attention map is applied to the feature map by the attention mechanism.
We use one of two types of attention mechanisms, i.e., Eq.~\ref{eq:nal} and Eq.~\ref{eq:arl}.
Here, $g_{c}({\bf x}_{i})$ is the feature map at the feature extractor, $M({\bf x}_{i})$ is an attention map, and $g'_{c}({\bf x}_{i})$ is the output of the attention mechanism, as shown in Fig.~\ref{fig:attebtion_branch_network}(a).
Note that $\{c|1,\ldots,C\}$ is the index of the channel.
\begin{eqnarray}
g'_{c}({\bf x}_{i}) &=& M({\bf x}_{i})\cdot g_{c}({\bf x}_{i}) \label{eq:nal}\\
g'_{c}({\bf x}_{i}) &=& (1+M({\bf x}_{i}))\cdot g_{c}({\bf x}_{i}) \label{eq:arl}
\end{eqnarray}
Equation~\ref{eq:nal} is simply a dot-product between the attention and feature maps at a specific channel~$c$.
In contrast, Eq.~\ref{eq:arl} can highlight the feature map at the peak of the attention map while preventing the lower value region of the attention map from degrading to zero.

\subsection{Training}	
ABN can be trainable in an end-to-end manner using losses at both branches.
Our training loss function~$L({\bf x}_{i})$ is a simple sum of losses at both branches, as expressed by Eq.~\ref{eq:abn_loss}.
\begin{eqnarray}
L({\bf x}_{i}) = L_{att}({\bf x}_{i}) + L_{per}({\bf x}_{i})
\label{eq:abn_loss}
\end{eqnarray}
Here, $L_{att}({\bf x}_{i})$ denotes training loss at the attention branch with an input sample~${\bf x}_{i}$, and $L_{per}({\bf x}_{i})$ denotes training loss at the perception branch.
Training loss for each branch is calculated by the combination of the softmax function and cross-entropy in image classification task.
The feature extractor is optimized by passing through the gradients of the attention and perception branches during back propagation.
If ABN is applied to other image recognition tasks, our training loss can adaptively change depending on the baseline model.

\subsection{ABN for multi-task learning}
ABN with a classification model outputs the attention map and final class probability by dividing the two branches.
This network design can be applicable to other image recognition tasks, such as multi-task learning.
In this section, we explain ABN for multi-task learning.

\begin{figure}[t]
\begin{center}
\includegraphics[width=1.0\linewidth]{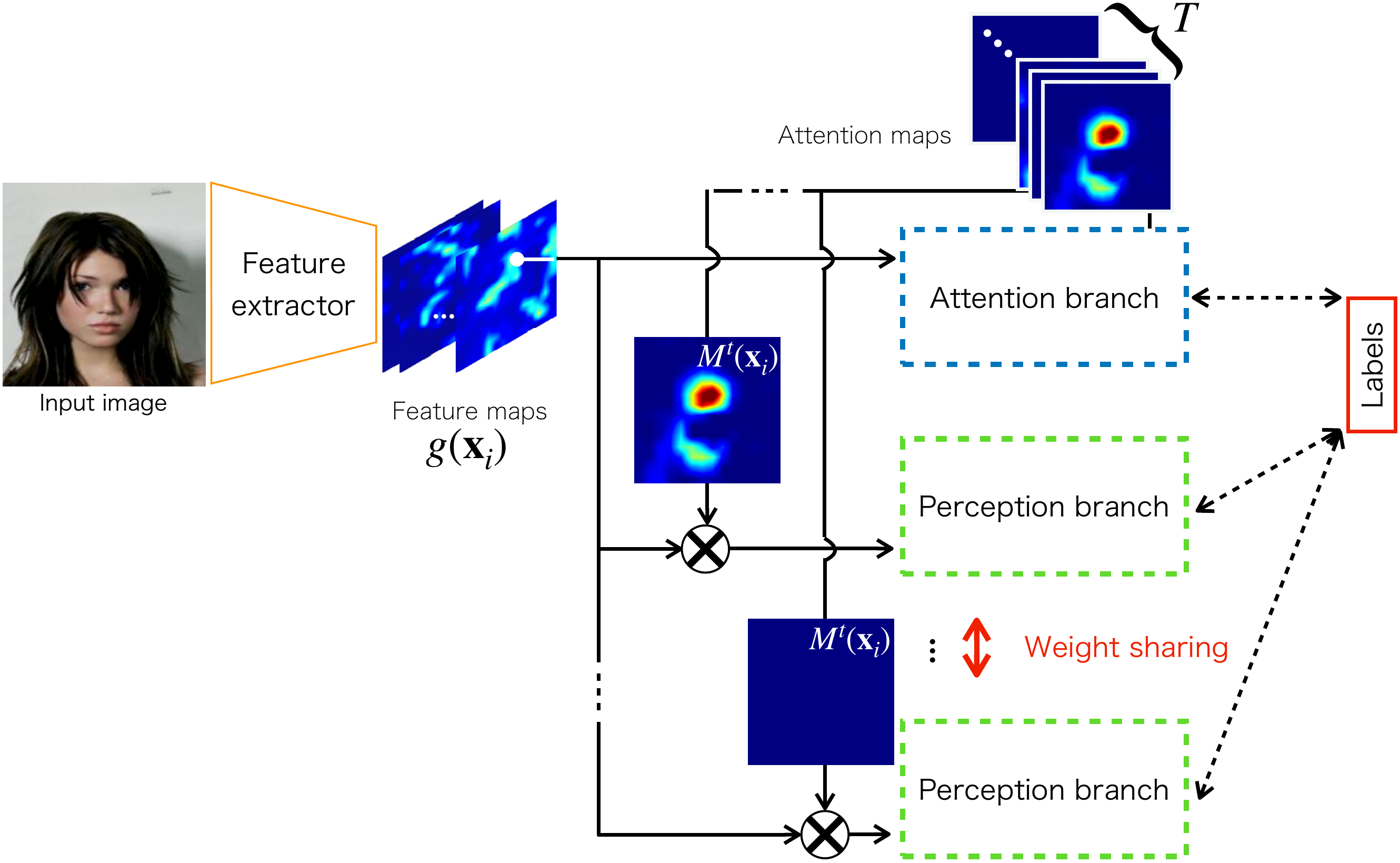}
\caption{ABN for multi-task learning.}
\label{fig:det_mtl}
\end{center}
\end{figure}

Conventional multi-task learning has units outputting the recognition scores corresponding to each task~\cite{Caruana1993}.
In training, the loss function defines multiple tasks using a single network.
However, there is a problem with ABN for multi-task learning.
In image classification, the relation between the numbers of inputs and recognition tasks is one-to-one.
In contrast, the relation between the numbers of inputs and recognition tasks of multi-task learning is one-to-many.
The one-to-one relation can be focused on the specific target location using a single attention map, but the one-to-many relation cannot be focused on multiple target locations using a single attention map.
To address this issue, we generate multiple attention maps for each task by introducing multi-task learning to the attention and perception branches.
Note that we use ResNet with multi-task learning as the baseline model.

To output multiple attention maps, we design the attention branch with multi-task learning, as shown in Fig.~\ref{fig:det_mtl}.
First, a feature map at residual block 4 is convoluted by the $T\times$1$\times$1 convolution layer, and the $T\times$14$\times$14 feature map is output.
The probability score during a specific task $\{t|1,\ldots, T\}$ is output by applying the 14$\times$14 feature map at specific task $t$ to GAP and the sigmoid function.
In training, we calculated the training loss by combining the sigmoid function and binary cross-entropy loss function.
We apply the 14$\times$14 feature maps to the attention maps.

We introduce the perception branch to multi-task learning.
Converting feature map~$g'^{t}_{c}({\bf x})$ is first generated using attention map~$M^{t}({\bf x})$ at specific task $t$ and feature map~$g({\bf x})$ at the feature extractor, as shown in Eq.~\ref{eq:per_branch_mtl_fea} in Sec.~\ref{sec:perception}.
After generating feature map~$g'^{t}_{c}({\bf x})$, the probability score at specific task $t$ is calculated on perception branch~$p_{per}(\cdot)$, which outputs the probability for each task by inputting feature map~$g'^{t}({\bf x})$.
\begin{eqnarray}
g'^{t}_{c}({\bf x}_{i}) &=& M^{t}({\bf x}_{i})\cdot g_{c}({\bf x}_{i}) \label{eq:per_branch_mtl_fea}\\
{\bf O}(g'^{t}_{c}({\bf x}_{i})) &=& p_{per}(g'^{t}_{c}({\bf x}_{i})~;~\theta) \label{eq:per_branch_mtl}
\end{eqnarray}
This probability matrix of each task~${\bf O}(g'^{t}_{c}({\bf x}_{i}))$ on the perception branch consists of $T\times2$ components defined two categories classification for each task.
The probability~${\bf O}^{t}(g'^{t}_{c}({\bf x}_{i}))$ at specific task $t$ is used when the perception branch receives the feature map~$g'^{t}_{c}({\bf x})$ that applies the attention map at specific task $t$, as shown in Fig.~\ref{fig:det_mtl}.
These processes are repeated for each task.

\section{Experiments}\label{sec:eval}

\subsection{Experimental details on image classification}\label{sec:ablation}
First, we evaluate ABN for an image classification task using the CIFAR10, CIFAR100, Street View Home Number~(SVHN)~\cite{Netzer2011}, and ImageNet~\cite{Deng2009} datasets.
The input image size of the CIFAR10, CIFAR100, SVHN datasets is 32$\times$32 pixels, and that of ImageNet is 224$\times$224 pixels.
The number of categories for each dataset is as follows: CIFAR10 and SVHN consist of 10 classes, CIFAR100 consists of 100 classes, and ImageNet consists of 1,000 classes.
During training, we applied the standard data augmentation. 
For CIFAR10, CIFAR100, and SVHN, the images are first zero-padded with 4 pixels for each side then randomly cropped to again produce 32$\times$32 pixels images, and the images are then horizontally mirrored at random.
For ImageNet, the images are resized to 256$\times$256 pixels then randomly cropped to again produce 224$\times$224 pixels images, and the images are then horizontally mirrored at random.
The numbers of training, validation, and testing images of each dataset are as follows: CIFAR10 and CIFAR100 consist of 60,000 training images and 10,000 testing images, SVHN consists of 604,388 training images (train:73,257, extra:531,131) and 26,032 testing images, and ImageNet consists of 1,281,167 training images and 50,000 validation images.

We optimize the networks by stochastic gradient descent~(SGD) with momentum.
On CIFAR10 and CIFAR100, the total number of iterations to update the parameters is 300~epochs, and the batch size is 256.
The total numbers of iterations to update the networks is as follows: CIFAR10 and CIFAR100 are 300 epochs, SVHN is 40 epochs, and ImageNet is 90 epochs.
The initial learning rate is set to 0.1, and is divided by 10 at 50~$\%$ and 75~$\%$ of the total number of training epochs.

\subsection{Image classification}

\begin{table}[t]
 \caption{Comparison of the top-1 errors on CIFAR100 with attention mechanism.}
 \label{table:attention_map_comp}
 \centering
  \scalebox{0.95}[0.95]{ 
  \begin{tabular}{c|c|c|c}
   \hline
  & $g({\bf x})$ & $g({\bf x})\cdot M({\bf x})$ & $g({\bf x})\cdot (1 + M({\bf x}))$ \\ \hline \hline
ResNet20 & 31.47 & 30.61 & {\bf 30.46} \\
ResNet32 & 30.13 & 28.34 & {\bf 27.91} \\
ResNet44 & 25.90 & {\bf 24.83} & 25.59 \\
ResNet56 & 25.61 & 24.22 & {\bf 24.07} \\
ResNet110 & 24.14 & 23.28 & {\bf 22.82} \\ 
   \hline
  \end{tabular}
    }
\end{table}

{\tabcolsep=0.08\linewidth
\begin{table*}[th]
\caption{Comparison of top-1 errors on CIFAR10, CIFAR100, SVHN, and ImageNet dataset.}
\label{table:comp_top1_error}
\begin{center}
\scalebox{1.0}[1.0]{ 
\begin{tabular}{c||c|c|c||c}
\hline
Dataset         & CIFAR10 & CIFAR100 & SVHN~\cite{Netzer2011}   & ImageNet~\cite{Deng2009} \\ \hline
%method         & original & re- & original  & re- & original & re- & original & re- \\ \hline\hline
% AlexNet~\cite{Alex2014}		& -- 				& -- 				& --					&42.6	\\
VGGNet~\cite{Simonyan2014}		& --				& -- 				& --					&31.2	\\
VGGNet+BN					& --				& -- 				& --					&$26.24^\ast$	\\
ResNet~\cite{He2016}			&6.43			&$24.14^\ast$		&$2.18^\ast$			&$22.19^\ast$	\\ \hline
% AlexNet+CAM~\cite{Zhou2016}	& --				& -- 				& --					&44.9	\\
VGGNet+CAM~\cite{Zhou2016}	& --				& -- 				& --					&33.4	\\
VGGNet+BN+CAM				& --				& -- 				& -- 					&$27.42^\ast_{~(+1.18)}$	\\
ResNet+CAM					& --				& -- 				& -- 					&$22.11^\ast_{(-0.08)}$	\\ \hline
WideResNet~\cite{Zagoruyko2016}	&4.00			&19.25			&$2.42^\ast$			&21.9	\\
DenseNet~\cite{Gao2017}			&4.51			& 22.27			&$2.07^\ast$			&22.2	\\
ResNeXt~\cite{Xie2017}			&$3.84^\ast$		&$18.32^\ast$		&$2.16^\ast$			&22.4	\\
Attention~\cite{Wang2017a}		&3.90			&20.45			& --					&21.76	\\
AttentionNeXt~\cite{Wang2017a}	& --				& --				& --					&21.20	\\
SENet~\cite{Hu2017}			& --				& --				& --					&21.57	\\ \hline
% AlexNet+ABN				& --				& --				& --					& --		\\
VGGNet+BN+ABN				& --				& --				& --					&$25.55_{~(-0.69)}$	\\
ResNet+ABN					&$4.91_{~(-1.52)}$	&$22.82_{~(-1.32)}$	&${\bf 1.86_{~(-0.32)}}$	&$21.37_{~(-0.82)}$	\\
WideResNet+ABN				&${\bf 3.78_{~(-0.22)}}$	&$18.12_{~(-1.13)}$	&$2.24_{~(-0.18)}$	& --				\\
DenseNet+ABN				&$4.17_{~(-0.34)}$	&$21.63_{~(-0.64)}$	&$2.01_{~(-0.06)}$		& --				\\
ResNeXt+ABN					&$3.80_{~(-0.04)}$	&${\bf 17.70_{~(-0.62)}}$	&$2.01_{~(-0.15)}$	& --				\\
SENet+ABN					& --				& --				& --					&${\bf 20.77_{~(-0.80)}}$	\\ \hline
\multicolumn{5}{r}{\small\it $\ast$ indicates results of re-implementation accuracy}\\
\end{tabular}
}
\end{center}
\end{table*}
}

{\bf Analysis on attention mechanism}\hspace{2mm}
We compare the accuracies of attention mechanisms Eq.~\ref{eq:nal} and Eq.~\ref{eq:arl}.
We use ResNet \{20, 33, 44, 56, 110\} models on CIFAR100.

Table~\ref{table:attention_map_comp} shows the top-1 errors of attention mechanisms Eq.~\ref{eq:nal} and Eq.~\ref{eq:arl}.
The $g({\bf x})$ is conventional ResNet.
First, we compare ABN with $g({\bf x})\cdot M({\bf x})$ attention mechanism at Eq.~\ref{eq:nal} and conventional ResNet~$g({\bf x})$.
Attention mechanism~$g({\bf x})\cdot M({\bf x})$ has suppressed the top-1 errors than conventional ResNet.
We also compare the accuracy of both $g({\bf x})\cdot M({\bf x})$ and $g({\bf x})\cdot (1 + M({\bf x}))$ attention mechanisms.
Attention mechanism~$g({\bf x})\cdot (1 + M({\bf x}))$ is slightly more accurate than attention mechanism~$g({\bf x})\cdot M({\bf x})$.
In residual attention network, which includes the same attention mechanisms, accuracy decreased with attention mechanism~$g({\bf x})\cdot M({\bf x})$~\cite{Wang2017a}.
Therefore, our attention map responds to the effective region in image classification.
We use attention mechanism~$g({\bf x})\cdot (1 + M({\bf x}))$ at Eq.~\ref{eq:arl} as default manner.

{\bf Accuracy on CIFAR and SVHN}\hspace{2mm}
Table~\ref{table:comp_top1_error} shows the top-1 errors on CIFAR10/100, SVHN, and ImageNet.
We evaluate these top-1 errors using various baseline models, CAM, and ABN regarding image classification.
These errors are an original top-1 error at referring paper~\cite{Simonyan2014,He2016,Zhou2016,Zagoruyko2016,Gao2017,Xie2017,Wang2017a,Wang2017a,Hu2017} or top-1 errors of our model, and the '$\ast$' indicates the results of re-implementation accuracy.
The numbers in brackets denote the difference in the top-1 errors from the conventional models at re-implementation.
On CIFAR and SVHN, we evaluate the top-1 errors by using the following ResNet models as follows: ResNet~(depth=110), DenseNet~(depth=100, growth rate=12), Wide ResNet~(depth=28, widen factor=4, drop ratio=0.3), ResNeXt~(depth=28, cardinality=8, widen factor=4).
Note that ABN is constructed by dividing a ResNet model at residual block 3.

Accuracies of ResNet, Wide ResNet, DenseNet and ResNeXt are improved by introducing ABN.
On CIFAR10, ResNet and DenseNet with ABNs decrease the top-1 errors from 6.43~$\%$ to 4.91~$\%$ and 4.51~$\%$ to 4.17~$\%$, respectively.
Additionally, all ResNet models are decrease the top-1 errors by more 0.6~$\%$ on CIFAR100.

{\bf Accuracy on ImageNet}\hspace{2mm}
We evaluate the image classification accuracy on ImageNet as shown in Table~\ref{table:comp_top1_error} in the same manner as for CIFAR10/100 and SVHN.
On ImageNet, we evaluate the top-1 errors by using the VGGNet~(depth=16), ResNet~(depth=152), and SENet~(ResNet152 model).
First, we compare the top-1 errors of CAM.
The performance of CAM slightly decreased with a specific baseline model because of the removal of the fully-connected layers and adding a GAP~\cite{Zhou2016}.
Similarly, the performance on VGGNet+BatchNormalization~(BN)~\cite{Sergey2015} with CAM decreases even in re-implementation.
In contrast, the performance of ResNet with CAM is almost the same as that of baseline ResNet.
The structure of the ResNet model that contains GAP and a fully-connected layer as the last layer resembles that in CAM.
ResNet with CAM can be easily constructed by stacking on the $K \times 1 \times 1$ convolution layer at the last residual block, which sets the stride to 1 at the first convolution layer.
Therefore, it is difficult to decrease the performance of ResNet with CAM due to removal of the fully-connected layer and adding GAP.
On the other hand, ABN outperforms conventional VGGNet and CAM and performs better than conventional ResNet and CAM.

We compare the accuracy of a conventional attention models.
By introducing the SE modules to ResNet152, SENet reduces the top-1 errors from 22.19$\%$ to 21.90$\%$.
However, ABN reduces the top-1 errors from 22.19~$\%$ to 21.37$\%$, indicating that ABN is more accurate than SENet.
Moreover, ABN can introduce the SENet in parallel.
SENet with ABN reduces the top-1 errors from 22.19~$\%$ to 20.77~$\%$ compared to the ResNet152.
Residual attention network results in the same amount of top-1 errors from the size of the input image, which is $224\times224$, as follows: ResNet is 21.76$\%$, and ResNeXt is 21.20$\%$.
Therefore, ResNet152+SENet with ABN indicates more accurate than these residual attention network models.

\begin{figure}[t]
\begin{center}
\includegraphics[width=1.0\linewidth]{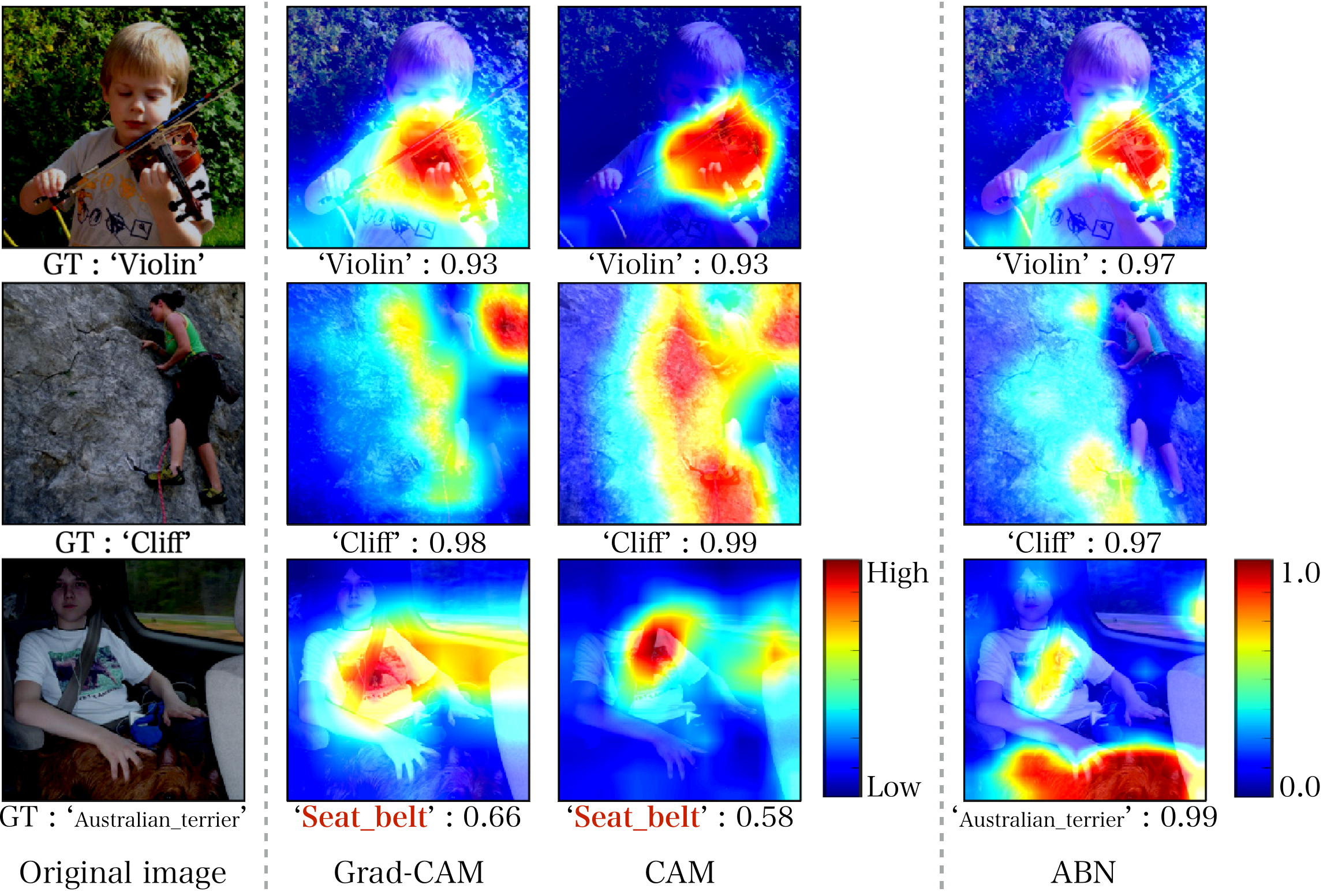}
\caption{Visualizing high attention area with CAM, Grad-CAM, and our ABN. CAM and Grad-CAM are visualized attention maps at top-1 result.}
\label{fig:visualizing_attentionmap}
\end{center}
\end{figure}

{\bf Visualizing attention maps}\hspace{2mm}
We compare the attention maps visualized using Grad-CAM, CAM, and ABN.
Grad-CAM generates an attention map by using ResNet152 as a baseline model.
CAM and ABN are constructed using ResNet152 as a baseline model.
Figure.~\ref{fig:visualizing_attentionmap} shows the attention maps for each model on ImageNet dataset.

This Fig.~\ref{fig:visualizing_attentionmap} shows that Grad-CAM, CAM and ABN highlights a similar region.
For example in the first column in Fig.~\ref{fig:visualizing_attentionmap}, these models classify the ``Violin", and highlight the ``Violin" region on the original image.
Similarly, they classify ``Cliff" in the second column and highlight the ``Cliff" region.
For the third column, this original image is a typical example because multiple objects, such as ``Seat belt" and ``Australian terrier", are included. 
In this case, Grad-CAM~(conventional ResNet152) and CAM failes, but ABN performs well.
When visualizing the attention maps in the third column, the attention map of ABN highlights each object.
Therefore, this attention map can focus on a specific region when multiple objects are in an image.

\begin{table}[t]
   \caption{Comparison of car model and maker accuracy on CompCars dataset}
   \label{tab:carcomp_accuracy}
\begin{center}
\begin{tabular}{c|c|c} \hline
task & ~model~[$\%$]~ & ~maker~[$\%$]~ \\ \hline\hline
VGG16 & 85.9 & 90.4 \\ 
ResNet101 & 90.2 & 90.1 \\ \hline
VGG16+ABN & 90.7 & 92.9 \\ 
ResNet101+ABN & {\bf 97.1} & {\bf 98.1} \\ 
\hline
\end{tabular}
\end{center}
\end{table}

\begin{figure}[t]
\begin{center}
\includegraphics[width=1.0\linewidth]{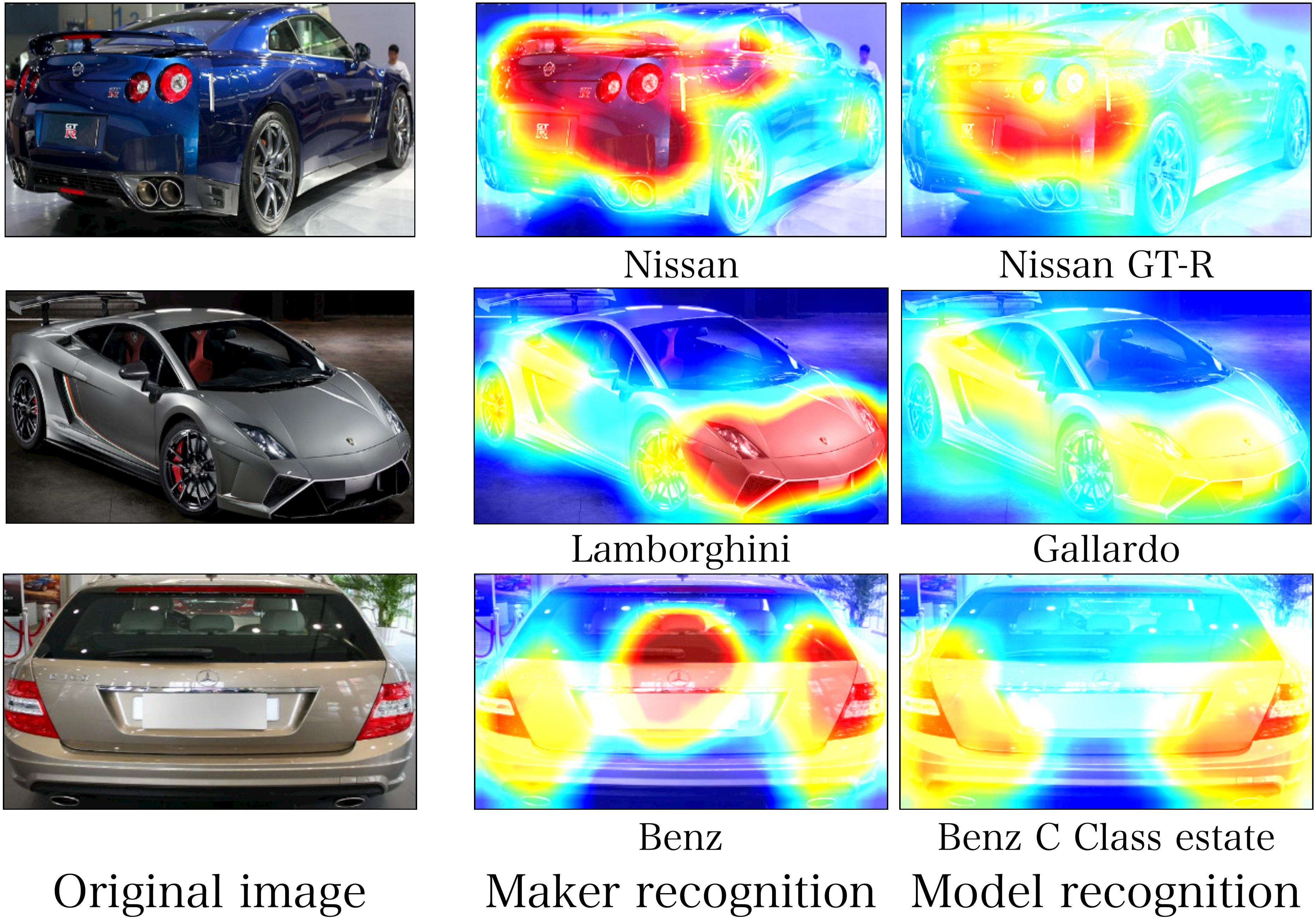}
\caption{Visualizing attention map on fine-grained recognition.}
\label{fig:attention_results_fine}
\end{center}
\end{figure}

\begin{figure}[t]
\begin{center}
\includegraphics[width=1.0\linewidth]{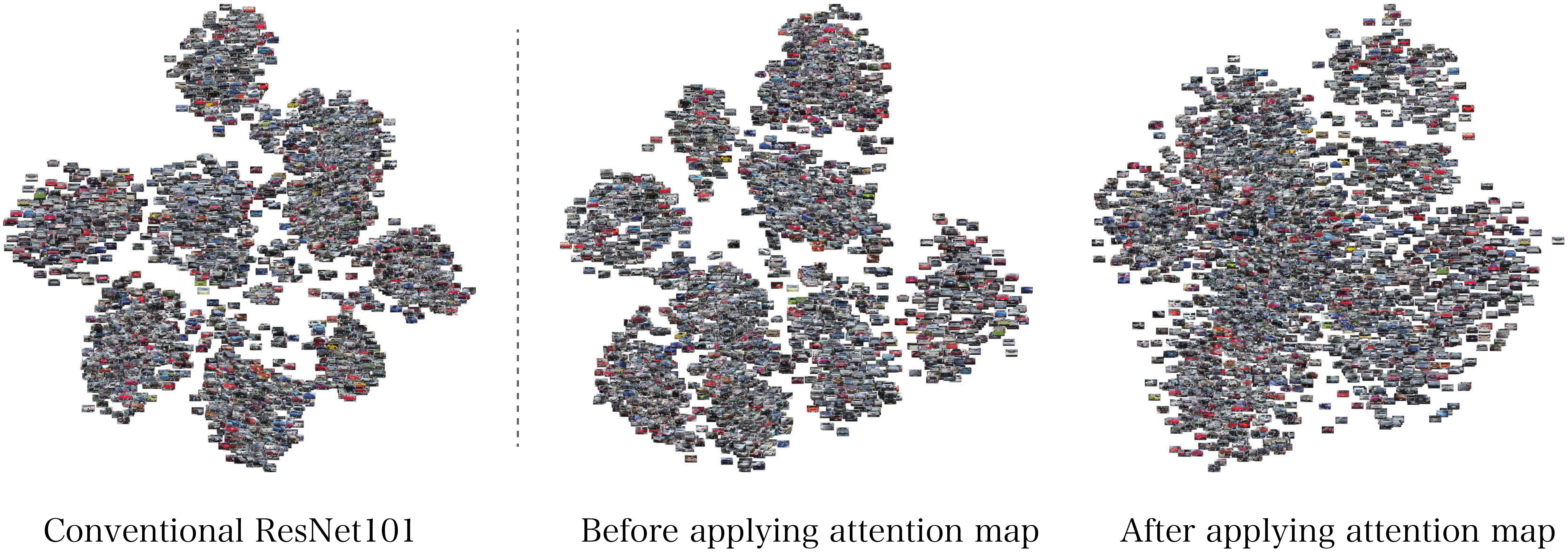}
\caption{Comparison of distribution maps at residual block 4 by t-SNE. {\bf Left : }distribution of baseline ResNet101 model. {\bf Center and Right : }distribution of ABN. {\bf Center} did not apply the attention map.}
\label{fig:comp_vis_feature}
\end{center}
\end{figure}

\begin{figure*}[t]
\begin{center}
\includegraphics[width=1.0\linewidth]{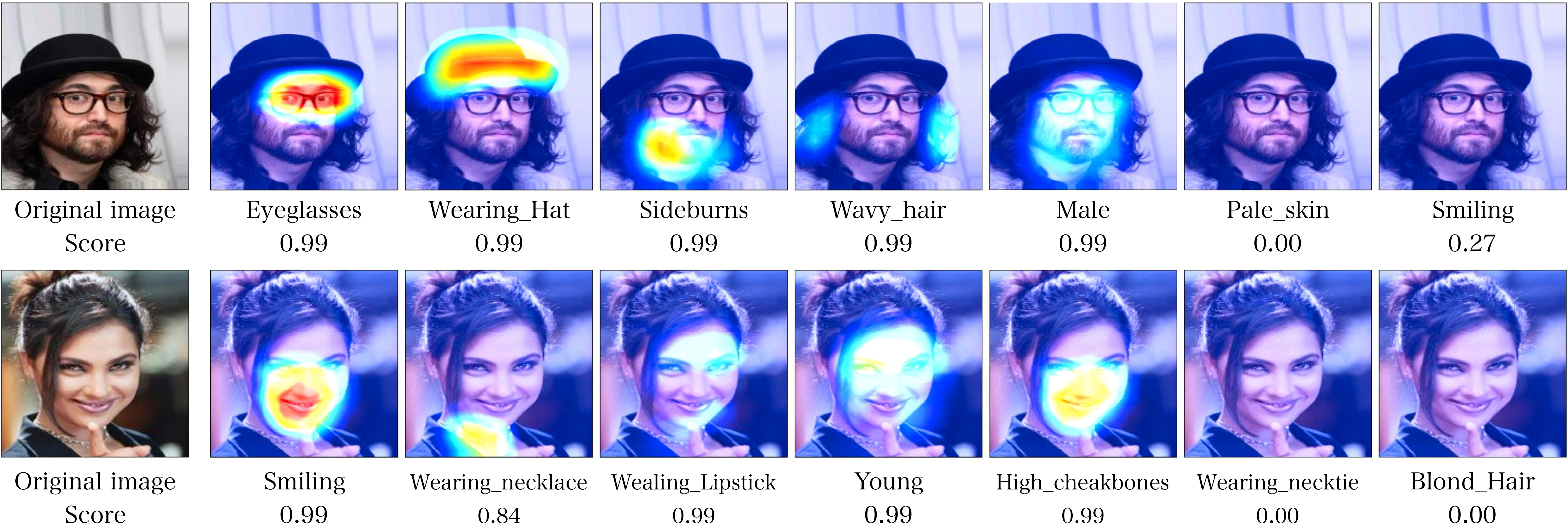}
\caption{Visualizing attention maps on multiple facial attributes recognition. These scores are final recognition scores at the perception branch.}
\label{fig:attention_results_img_recog}
\end{center}
\end{figure*}

\subsection{Fine-grained recognition}
We evaluate the performance of ABN for the fine-grained recognition on the comprehensive cars~(CompCars) dataset~\cite{Linjie2015}, which has 36,451 training images and 15,626 testing images with 432 car models and 75 makers.
We use VGG16 and ResNet101 as baseline model and optimized these models by SGD with momentum.
The total number of update iterations is 50~epochs, and the mini-batch size is 32.
The learning rate starts from 0.01 and is multiplied by 0.1 at 25 and 35 epochs.
The input image is resized to 323$\times$224 pixels.
The image size is calculated by taking the average of the bounding box aspect ration from the training data.
This resizing process prevents the collapse of the car shape.

Table~\ref{tab:carcomp_accuracy} shows the car model and maker recognition accuracy on the CompCars dataset.
The car model recognition accuracy of ABN improves by 4.9 and 6.2~$\%$ with VGG16 and ResNet101, respectively.
Moreover, maker recognition accuracy improves by 2.0 and 7.5~$\%$, respectively. 
These results indicate that ABN is also effective for fine-grained recognition. 
We visualize the attention maps for car model or maker recognition, as shown in Fig.~\ref{fig:attention_results_fine}.
From these visualizing results, training and testing images are the same for car model and maker recognition, however, our attention maps differ depending on the recognition task.

\begin{table}[t]
   \caption{Comparison of multiple facial attribute recognition accuracy on CelebA dataset}
   \label{tab:celeba_accuracy}
\begin{center}
\scalebox{1.0}{
\begin{tabular}{c|c|c} \hline
Method & Average of accuracy~[\%] & Odds \\ \hline
FaceTracer~\cite{Kumar2008} & 81.13 & 40/40 \\
PANDA-l~\cite{Ning2014} & 85.43 & 39/40 \\
LNet+ANet~\cite{Liu2015} & 87.30 & 37/40 \\
MOON~\cite{Ethan2016} & 90.93 & 29/40 \\ 
ResNet101 & 90.69 & 27/40 \\
ABN & {\bf 91.07} & -- \\
\hline
\end{tabular}
}
\end{center}
\end{table}

We compare the feature representations of the ResNet101 and ResNet101 with ABN.
We visualize distributions by t-distributed stochastic neighbor embedding~(t-SNE)~\cite{Van2008} and analyze the distributions.
We use the comparison feature maps at the final layer on residual block 4.
Figure~\ref{fig:comp_vis_feature} shows the distribution maps of t-SNE.
We use 5,000 testing images on the CompCars dataset.
The feature maps of ResNet101 and the feature extractor in the attention branch network are clustered by car pose.
However, the feature map applying the attention map is split distribution by car pose and detail car form.

\subsection{Multi-task Learning}
For multi-task learning, we evaluate for multiple facial attributes recognition using the CelebA dataset~\cite{Liu2015}, which consists of 202,599 images~(182,637 training images and 19,962 testing images) with 40 facial attribute labels.
The total number of iterations to update the parameters is 10~epochs, and the learning rate is set to 0.01.
We evaluate the accuracy rate using FaceTracer~\cite{Kumar2008}, PANDA-l~\cite{Ning2014}, LNet+ANet~\cite{Liu2015}, mixed objective optimization network~(MOON)~\cite{Ethan2016}, and ResNet101.

Table~\ref{tab:celeba_accuracy} shows that ABN outperforms all conventional methods regarding the average recognition rate and number of facial attribute tasks.
Note that the numbers in the third column in Table~\ref{tab:celeba_accuracy} are the numbers of winning tasks when we compare the conventional models with ABN for each facial attribute.
The accuracy of a specific facial attribute task is described in the appendix.
When we compare ResNet101 and ABN, ABN is 0.38$\%$ more accurate.
Moreover, the accuracy of 27 facial tasks is improved.
ABN also performs better than conventional facial attribute recognition models, i.e., FaceTracer, PANDA-l, LNet+ANet, MOON.
ABN outperforms these models for difficult tasks such as ``arched eyebrows", ``pointy nose", ``wearing earring", and ``wearing necklace".
Figure~\ref{fig:attention_results_img_recog} shows the attention map of ABN on CelebA dataset.
These attention maps highlights the specific locations such as  mouth, eyes, beard, and hair.
These highlight locations correspond to the specific facial task, as shown in Fig.~\ref{fig:attention_results_img_recog}.
It is conceivable that these highlight locations contributed to performance improvement of ABN.

\section{Conclusion}
We propose an Attention Branch Network, which extends a response-based visual explanation model by introducing a branch structure with an attention mechanism.
ABN can be simultaneously trainable for visual explanation and improving the performance of image recognition with an attention mechanism in an end-to-end manner.
It is also applicable to several CNN models and image recognition tasks.
We evaluated the accuracy of ABN for image classification, fine-grained recognition, and multi-task learning, and it outperforms conventional models for these tasks.
We plan to apply ABN to reinforcement learning that does not include labels in the training process.

{\small
\bibliographystyle{ieee}
\bibliography{egpaper_final}
}

\end{document}